\documentclass[10pt,leqno]{amsart}

\usepackage{amsmath,amssymb,amsfonts}
\usepackage{graphicx}
\usepackage{subcaption}
\usepackage{booktabs}
\usepackage{array}
\usepackage{url}
\usepackage{csquotes}
\usepackage{hyperref}
\usepackage{indentfirst}
\usepackage{chngcntr}

\hypersetup{ colorlinks=true, linkcolor=black, filecolor=black, urlcolor=black }

\makeatletter
\renewcommand{\subsection}{
  \@startsection{subsection}{2}{0pt}
    {1.5ex \@plus .5ex \@minus .2ex}
    {1.0ex \@plus .2ex}          
    {\normalfont\large\bfseries}}
\makeatother

\begin{document}
\title{Decoding Motor Behavior Using Deep Learning and Reservoir Computing}
\author{Tian Lan}
\address{Graduate School of Economics, Hitotsubashi University, Tokyo, Japan}
\footnotetext{This manuscript is based on the master’s thesis submitted by the author to Hitotsubashi University.}
\email{em245080@g.hit-u.ac.jp; lantian0329@gmail.com}
\maketitle

\begin{abstract}
We present a novel approach to EEG decoding for non-invasive brain–machine interfaces (BMIs), with a focus on motor-behavior classification. While conventional convolutional architectures such as EEGNet and DeepConvNet are effective in capturing local spatial patterns, they are markedly less suited for modeling long-range temporal dependencies and nonlinear dynamics \cite{ref1,ref2}. To address this limitation, we integrate an Echo State Network (ESN), a prominent paradigm in reservoir computing into the decoding pipeline \cite{ref3,ref4,ref5}. ESNs construct a high-dimensional, sparsely connected recurrent reservoir that excels at tracking temporal dynamics, thereby complementing the spatial representational power of CNNs. Evaluated on a skateboard-trick EEG dataset preprocessed via the PREP pipeline and implemented in MNE-Python, our ESNNet achieves 83.2\% within-subject and 51.3\% LOSO accuracies, surpassing widely used CNN-based baselines \cite{ref6,ref7}. Code is available at: \url{https://github.com/Yutiankunkun/Motion-Decoding-Using-Biosignals} 
\end{abstract} 

\section{Introduction}
Electroencephalogram (EEG) decoding constitutes a cornerstone of non-invasive BMIs, with potential applications in assistive technology, rehabilitation, and skill training. Prevailing EEG classification approaches typically follow a three-stage pipeline: preprocessing, feature extraction, and classification \cite{ref8}. Classical methods such as common spatial patterns (CSP), often combined with linear discriminant analysis (LDA) or support vector machines (SVMs), are computationally efficient and interpretable \cite{ref9,ref10}, yet fundamentally constrained by their reliance on static and linear assumptions. Such limitations render them ill-equipped to capture the high-dimensional, non-stationary, and dynamic characteristics of EEG signals, thereby reducing generalizability in real-world contexts.

The advent of deep learning has spurred CNN-based architectures, including EEGNet, DeepConvNet, and ShallowConvNet, which automatically extract spatial–spectral features in an end-to-end manner \cite{ref1,ref2,ref11}. These models achieve strong within-subject performance, yet their limited temporal receptive fields make them less effective at capturing long-range temporal structures. Hybrid methods such as CNN-LSTM frameworks partly remedy this issue, yet they introduce heavy training costs and susceptibility to overfitting \cite{ref12,ref13}, constraints that are particularly problematic in EEG scenarios characterized by high frequency and low signal-to-noise ratios.

Motivated by these gaps, we propose ESNNet, a hybrid model coupling CNN-based spatial–spectral extraction with an ESN module for temporal modeling. Unlike recurrent or Transformer-based architectures that require backpropagation through time, ESNs offer both stability and efficiency. This property is especially advantageous in real-time or resource-constrained BMI systems. In our design, raw EEG signals are preprocessed, passed through temporal and spatial convolution layers, and subsequently fed into an ESN reservoir, yielding a lightweight yet expressive temporal representation. To further enhance practical utility, we implement a GPU-accelerated ESN in PyTorch, enabling seamless integration with CNN modules \cite{ref23}.

Empirical evaluation on a skateboard-trick dataset demonstrates that ESNNet not only matches or outperforms state-of-the-art baselines, including EEGConformer, but also achieves favorable computational efficiency \cite{ref14}. Although all models face substantial accuracy degradation in cross-subject evaluations, a persistent challenge in EEG decoding, our framework exhibits comparatively robust generalization \cite{ref26}.

\section{Related Work}
\noindent\textbf{Handcrafted Features and Shallow Classifiers.}
Traditional EEG decoding has relied on engineered features, including CSP, wavelet transforms, $\mu/\beta$-band power, and autoregressive coefficients \cite{ref9,ref15,ref10}. These are typically paired with shallow classifiers such as LDA or SVMs, and occasionally random forests. Despite interpretability, such methods struggle with non-stationarity and high-dimensional temporal dynamics.

\noindent\textbf{Deep Neural Architectures.}
CNN-based models have since become dominant. EEGNet employs temporal convolutions; DeepConvNet and ShallowConvNet target high-frequency components; and FBCNet approximates filter-bank analysis \cite{ref1,ref2,ref16}. To capture temporal structure, hybrids such as CNN--LSTM, CTNet, and EEGConformer have been proposed \cite{ref12,ref14,ref19}. Yet, Transformer-based approaches remain data- and computation-intensive \cite{ref17,ref18,ref20}.

\noindent\textbf{Reservoir Computing Approaches.}
Reservoir computing offers an alternative that balances efficiency with expressive dynamics \cite{ref4,ref5}. ESNs, in particular, have shown strong potential for EEG classification and continue to be refined through modular and physical RC variants \cite{ref5,ref21,ref22}. Our work extends this line by coupling a CNN front-end with an ESN back-end, thereby exploiting both spatial--spectral and temporal representations in a computationally efficient framework.

\section{Methods}
\noindent\textbf{Pipeline.}
The overall pipeline of the ESNNet is illustrated in Fig.\,1. Given EEG signals and the corresponding action labels, mainstream EEG-based motion classification methods such as EEGNet typically perform decoding through convolutional neural architectures. We propose a hybrid architecture that integrates Convolutional Neural Networks (CNNs) with Echo State Networks (ESNs), aiming to capture temporal dependencies more effectively.

\begin{figure*}[t]
    \centering
    \includegraphics[width=\textwidth]{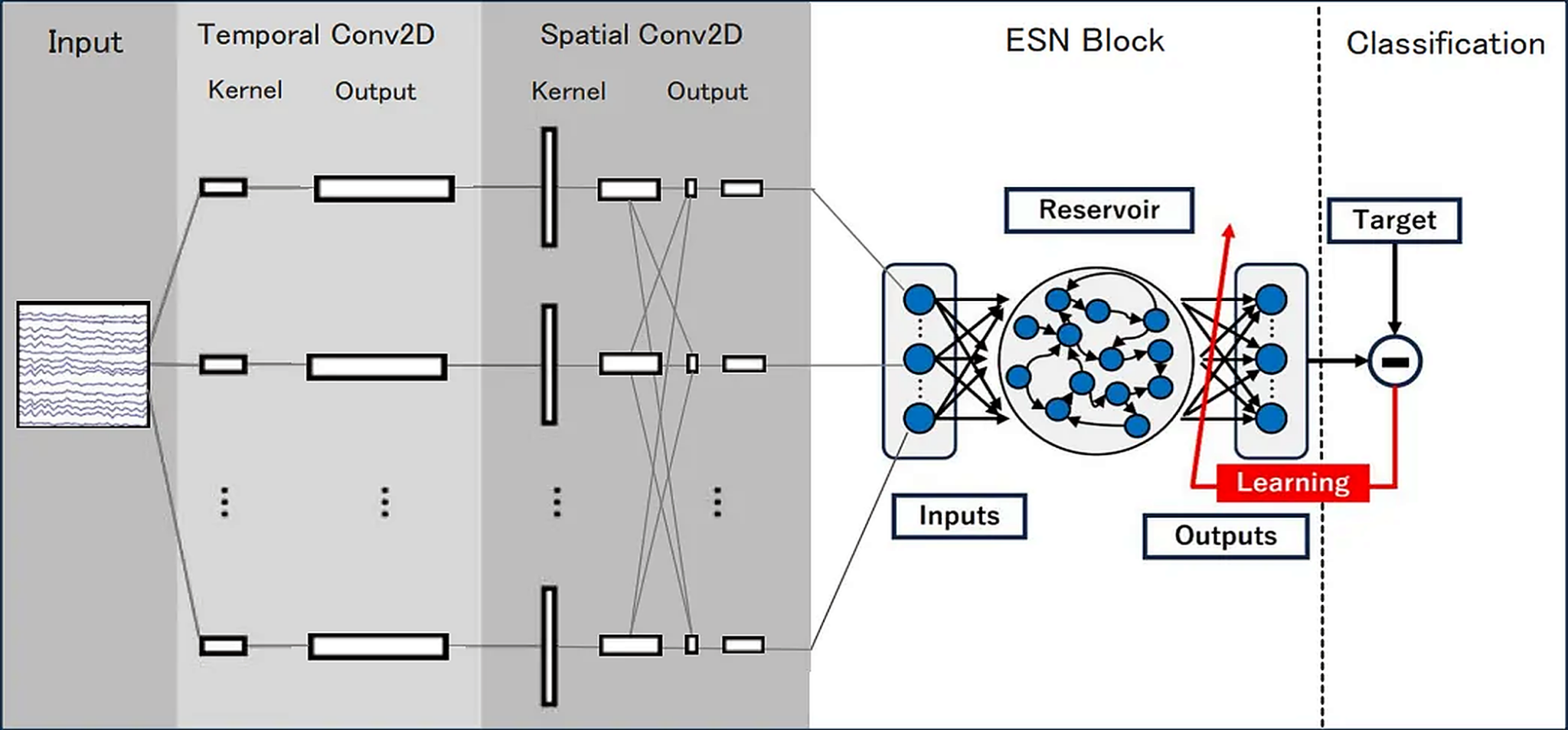}
    \caption{\textbf{Illustration of the architecture of the proposed model, which integrates convolutional feature extraction with an ESN for classification.}~
    The input time-series signals are first processed by a Temporal Conv2D layer to capture temporal dependencies, followed by a Spatial Conv2D layer that learns spatial correlations across channels. The resulting feature maps are then projected into the ESN block, where the reservoir dynamics enrich the temporal representation. The reservoir $W$ is fixed; the input mapping $W_{\text{in}}$, convolutional layers, and the linear classifier are trained jointly end-to-end, ensuring efficient learning. Finally, the ESN outputs are fed into a classifier, and the prediction is compared with the target labels to complete the supervised learning process.}
    \label{fig:preprocessing}
\end{figure*}

\subsection{EEG Feature Extraction}
To derive discriminative spatiotemporal features from raw EEG signals, we adopt the dual-convolutional structure introduced in EEGNet \cite{ref1}, employing a compact set of convolutional modules to separately process temporal and spatial information. Let the input signal be $U \in \mathbb{R}^{C \times T}$, where $C$ denotes the number of EEG channels and $T$ the sampling length.

\noindent\textbf{Temporal Convolution Layer.}
The temporal convolution layer $\text{Conv}_{\text{time}}$ applies a one-dimensional convolution independently to each channel, functioning as a set of learnable band-pass filters that extract oscillatory components across frequency bands:

\begin{equation}
    F_{\mathrm{temp}}
    =
    \phi\!\left(
        \operatorname{BN}\!\left(
            \operatorname{Conv}_{\mathrm{temp}}(U)
        \right)
    \right),
    \qquad
    F_{\mathrm{temp}} \in \mathbb{R}^{D \times C \times T},
\end{equation}

\noindent where BN denotes batch normalization and the kernel size is $1 \times k$, implemented as a temporal convolution. A nonlinear activation $\phi(\cdot)=\text{ELU}(\cdot)$ follows, producing the intermediate representation $F_{\text{temp}}$.

\noindent\textbf{Spatial Convolution Layer.}
To further capture inter-channel spatial dependencies, we apply a spatial convolution layer $\text{Conv}_{\text{spatial}}$ to $F_{\text{temp}}$ at each time step:

\begin{equation}
    F_{\mathrm{spatial}}
    =
    \phi\!\left(
        \operatorname{BN}\!\left(
            \operatorname{Conv}_{\mathrm{spatial}}(F_{\mathrm{temp}})
        \right)
    \right),
    \qquad
    F_{\mathrm{spatial}} \in \mathbb{R}^{D \times T'},
\end{equation}

\noindent where $T'$ denotes the temporal dimension after spatial convolution, and $T'=T$ because the kernel does not alter the temporal resolution. This operation, equivalent to a learnable version of the CSP filter, linearly combines EEG channels at each time point \cite{ref9}:

\begin{equation}
    f^{(d)}_t
    =
    \sum_{c=1}^{C}
        w^{(d)}_c \, x^{(c)}_t,
\end{equation}

\noindent where $x^{(c)}_t$ denotes the value of channel $c$ at time $t$ and $w^{(d)}_c$ the corresponding spatial weight for the $d$-th filter.

\subsection{ESN-based Modeling}
To strengthen temporal modeling capacity, we integrate an ESN module after feature extraction. Unlike conventional convolutional stacks or attention mechanisms, the ESN introduces a lightweight yet dynamically expressive reservoir structure, leveraging a fixed recurrent network and a trainable input mapping.

\noindent\textbf{Preliminaries: Echo State Network.}
The ESN, a canonical instantiation of Reservoir Computing, comprises a sparsely connected, randomly initialized recurrent reservoir with state-update dynamics:
\[
h_{t+1} = (1-\alpha)h_t + \alpha f\!\left(W h_t + W_{\text{in}} u_t\right)
,\]
where $f(\cdot)=\tanh(\cdot)$, $h_t \in \mathbb{R}^H$ denotes the reservoir state, $u_t$ the input, $W$ the recurrent weights, and $W_{\text{in}}$ the input weights. The leak rate $\alpha \in (0,1]$ controls the inertia of state updates. 
In our implementation, the reservoir $W$ is fixed, while the input mapping $W_{\text{in}}$, the convolutional front-end, and the linear classifier are learned jointly in an end-to-end manner.

\noindent\textbf{Training Strategy.}
The ESN reservoir $W$ remains fixed. 
The input mapping $W_{\text{in}}$, the convolutional feature extractor, and the linear classifier are optimized jointly end-to-end using the cross-entropy objective with L2 regularization. 
This keeps the optimization pipeline consistent with the softmax-based classifier and avoids mixing a closed-form ridge solution with a cross-entropy objective.

\noindent\textbf{Advantages and Limitations.}
A key advantage is that only the reservoir is fixed, ensuring few parameters, fast convergence, and robustness against overfitting. The random nonlinear reservoir facilitates efficient capture of complex temporal dependencies. Limitations include that fixed reservoir weights constrain expressiveness; performance is sensitive to hyperparameters such as $\rho$, $\alpha$, and reservoir size.

\subsection{Classification Head}
The output representations are aggregated through Global Average Pooling (GAP), followed by a fully connected layer that produces three-class logits. A softmax layer subsequently generates the final predictions. This architecture enhances robustness against temporal shifts and noise perturbations.

\subsection{Training Objective}
The training objective combines cross-entropy loss with $L_2$ regularization:

\[
\mathcal{L}(\theta)
= -\frac{1}{N}\sum_{i=1}^{N}\sum_{c=1}^{C}
y_{ic}\,\log\!\big([\mathrm{softmax}(\mathbf z_i)]_{c}\big)
\;+\; \lambda \lVert \theta \rVert_{2}^{2}
,\]

\noindent where $z_{i}$ denotes the logits produced by the linear classifier for the $i$-th sample, $\theta$ denotes all trainable parameters, $\lambda$ is the regularization coefficient, and $y_{ic}$ the ground-truth label.

\subsection{Complexity and Implementation}
We benchmark ESNNet against baseline models in terms of parameter count, floating-point operations (FLOPs), and inference latency measured on NVIDIA RTX 3090. Results (Table 2) indicate that ESNNet achieves competitive accuracy while reducing computational overhead. Specifically, ESNNet comprises approximately 46k parameters, 5.1M FLOPs per sample, and an inference latency of 0.6 ms—outperforming state-of-the-art models such as DeepConvNet and EEGConformer in efficiency \cite{ref2,ref14}.

\subsection{Implementation Details}
All models are trained and evaluated on the skateboard EEG dataset using the Adam optimizer with an initial learning rate of $1\times10^{-3}$ and batch size of 64. Training is conducted on a single RTX 3090 GPU, with ESNNet converging in roughly three hours. The feature extraction backbone follows EEGNet’s dual-convolutional design, coupled with a custom ESN module ($H=100,\,\rho=0.99,\,\alpha=0.1$), where H and $\rho$ denote the reservoir size of the ESN and the spectral radius, and $\alpha$ is the leak rate.

Input signals are sampled at 500 Hz across 72 channels, preprocessed via band-pass filtering ($1\sim40$ Hz) and z-score normalization. Each sample is a fixed-length EEG segment centered on event markers. During inference, ESNNet processes a single EEG segment within about 0.6 ms, making it well-suited for deployment in resource-constrained environments.

\section{Experiments}

\subsection{Dataset and Preprocessing}\label{sec:4.1}
\noindent\textbf{Dataset.}
We evaluate our approach on the publicly available skateboard EEG dataset released by the New Energy and Industrial Technology Development Organization (NEDO) \cite{ref24}. The dataset comprises recordings from five subjects performing three real-world skateboarding maneuvers: backside kickturn, frontside kickturn, and pumping. Each trial includes both the movement label and onset time annotation. EEG was recorded from 72 channels at a sampling rate of 500 Hz.

\noindent\textbf{Preprocessing.}
For each annotated trial, we extract a 500 ms segment beginning 0.2--0.7 seconds post-onset, resulting in samples of dimension $72\times250$. All data are standardized into the MNE RawArray format, with electrode positions mapped to the 10--20 system. Input signals undergo z-score normalization across channels. During training, data augmentation strategies are applied, including additive Gaussian noise ($\sigma=0.01$), temporal shifting, and signal inversion. These augmentations are stochastically applied each epoch, while the validation set remains unaltered.

Fig.\,2(a) provides an example of the raw EEG recordings across 72 channels, illustrating the signal complexity and the necessity of preprocessing. Fig.\,2(b) shows the temporal distribution of annotated events across the three maneuvers and two lighting conditions, ensuring transparency about trial segmentation and event balance.

\begin{figure*}[!t]
    \centering
    \begin{minipage}[b]{0.45\linewidth}
        \centering
        \includegraphics[width=\linewidth]{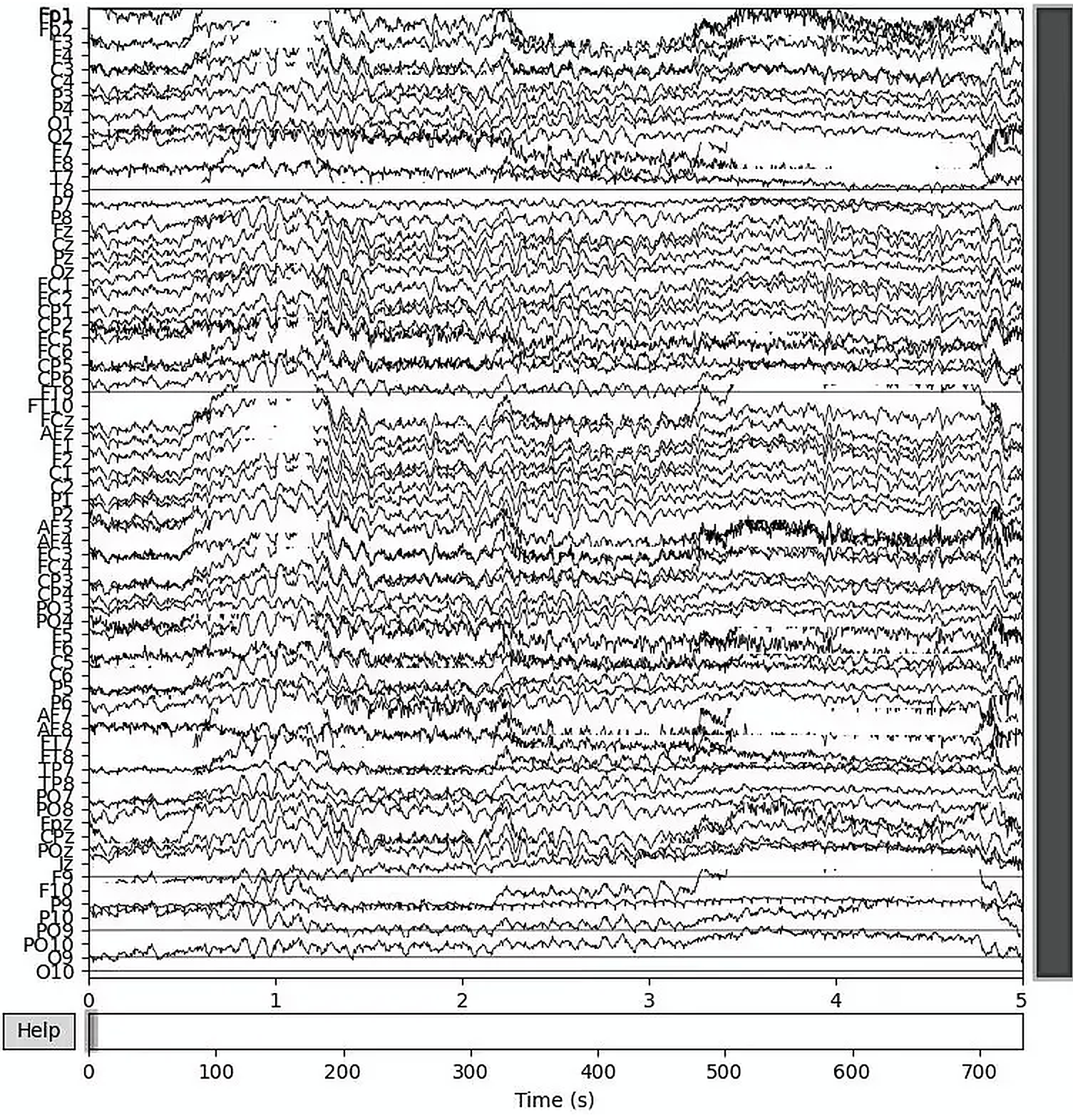}
    \end{minipage}
    \hfill
    \begin{minipage}[b]{0.45\linewidth}
        \centering
        \includegraphics[width=\linewidth]{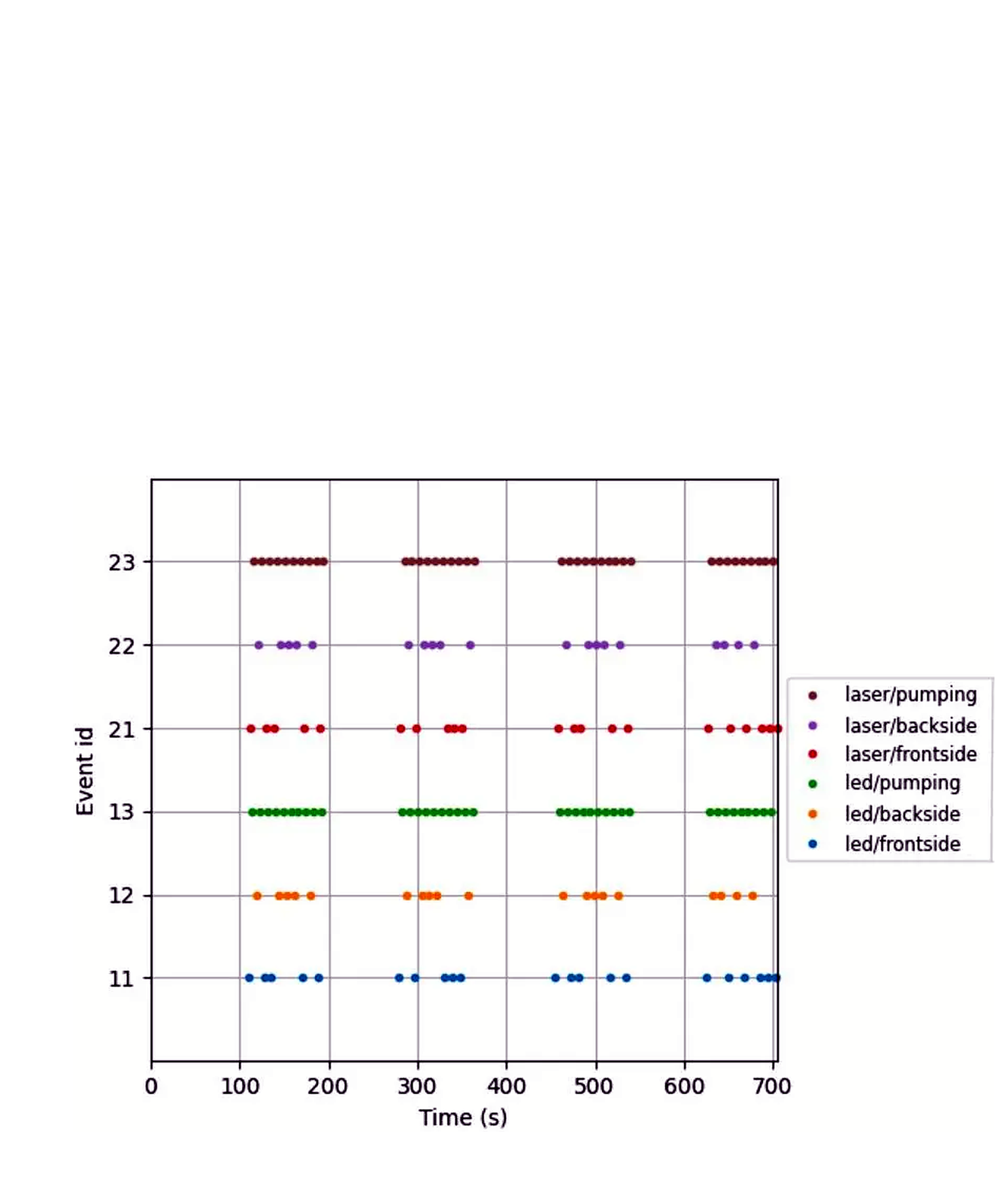}
    \end{minipage}
    \caption{\textbf{Overview of raw EEG data and events.}~
    \textbf{(a)}\,illustration of the temporal structure and variability of neural activity prior to preprocessing. The raw signals highlight characteristic fluctuations, background noise, and inter-channel differences, thereby motivating the subsequent preprocessing procedures described in subsection\,\ref{sec:4.1}. \textbf{(b)}\,Temporal distribution of labeled events for three skateboarding maneuvers (backside kickturn, frontside kickturn, and pumping) across two conditions with laser and LED. Each dot corresponds to an onset annotation, providing transparency on trial segmentation and ensuring balanced coverage of experimental factors.}
    \label{fig:preprocessing}
\end{figure*}

\subsection{Within-Subject Classification}
\noindent\textbf{Evaluation.}
Experiments are repeated under five random seeds, reporting mean accuracy and standard deviation. For each subject, trials are stratified by class and split into training and validation sets with a 7:3 ratio. Data augmentations identical to those described in preprocessing are applied during training. All models are optimized with Adam, learning rate = 0.001, up to 40 epochs, with early stopping employed to mitigate overfitting.

\noindent\textbf{Results.} 
As summarized in Table\,\ref{tab:1}, all models significantly outperform random chance, demonstrating the discriminability of EEG signals associated with skateboarding actions \cite{ref25}. ESNNet and DeepConvNet achieve the best performance, with ESNNet reaching an average validation accuracy of 83.3\% $\pm$ 1.7\%, compared with 80.2\% $\pm$ 1.9\% for DeepConvNet. ESNNet further exhibits lower variance across repeated trials, indicating enhanced training stability.

As summarized in Table\,\ref{tab:2}, pumping is consistently the most identifiable class, with ESNNet achieving 0.88 and EEGNet 0.87. By contrast, frontside and backside maneuvers exhibit pronounced confusion, with many models reporting F1-scores below 0.75. Notably, ESNNet maintains superior balance across all categories, particularly excelling in distinguishing between the confusable kickturn classes.

These findings indicate that ESNNet not only yields higher aggregate accuracy but also achieves more equitable class-level performance, reinforcing its robustness in fine-grained behavioral decoding.

\begin{table}[htbp]
  \centering
  \setlength{\tabcolsep}{0.01pt}
  \renewcommand{\arraystretch}{1.2}
  \begin{tabular}{lcccccc}
    \toprule
    Subject & EEGNet & DeepConvNet & ShallowConvNet & FBCNet & EEGConformer & {\bfseries\boldmath ESNNet} \\
    \midrule
    Subject 0 & \(79.9 \pm 2.7\)\% & \(81.6 \pm 1.8\)\% & \(70.8 \pm 2.9\)\% & \(78.8 \pm 2.2\)\% & \(81.1 \pm 2.4\)\% & {\bfseries\boldmath \(83.3 \pm 1.7\)\%} \\
    Subject 1 & \(78.1 \pm 2.5\)\% & \(80.8 \pm 2.0\)\% & \(69.3 \pm 3.0\)\% & \(76.5 \pm 2.6\)\% & \(79.4 \pm 2.2\)\% & {\bfseries\boldmath \(82.7 \pm 1.6\)\%} \\
    Subject 2 & \(76.3 \pm 3.1\)\% & \(78.5 \pm 1.5\)\% & \(67.1 \pm 3.3\)\% & \(73.9 \pm 2.9\)\% & \(75.6 \pm 2.8\)\% & {\bfseries\boldmath \(79.5 \pm 2.0\)\%} \\
    Subject 3 & \(87.5 \pm 1.6\)\% & \(86.2 \pm 1.9\)\% & \(74.6 \pm 2.5\)\% & \(81.3 \pm 2.1\)\% & \(85.1 \pm 2.0\)\% & {\bfseries\boldmath \(88.6 \pm 1.4\)\%} \\
    Subject 4 & \(77.7 \pm 2.4\)\% & \(73.2 \pm 2.7\)\% & \(65.9 \pm 2.8\)\% & \(72.0 \pm 2.5\)\% & \(78.3 \pm 2.3\)\% & {\bfseries\boldmath \(82.0 \pm 1.5\)\%} \\
    Macro Average & \(79.9 \pm 4.4\)\% & \(80.1 \pm 4.7\)\% & \(69.5 \pm 3.4\)\% & \(76.5 \pm 3.7\)\% & \(79.9 \pm 3.5\)\% & {\bfseries\boldmath \(83.2 \pm 3.3\)\%} \\
    \bottomrule
  \end{tabular}
  \caption{
  \textbf{Within-subject evaluation on classification across subjects.}
  Each entry reports mean accuracy $\pm$ standard deviation (std) over five independent runs.
  The ESNNet column is highlighted in bold.
  }
  \label{tab:1}
\end{table}

\begin{table}[htbp]
  \centering
  \setlength{\tabcolsep}{4pt}
  \renewcommand{\arraystretch}{1.25}
  \begin{tabular}{lccc}
    \toprule
    Model & Backside (F1) & Frontside (F1) & Pumping (F1) \\
    \midrule
    EEGNet       & 0.67 & 0.75 & 0.87 \\
    DeepConvNet      & 0.74 & 0.85 & 0.85 \\
    ShallowConvNet   & 0.66 & 0.67 & 0.84 \\
    FBCNet       & 0.59 & 0.48 & 0.78 \\
    EEGConformer & 0.68 & 0.61 & 0.83 \\
    \textbf{ESNNet}   & \textbf{0.70} & \textbf{0.75} & \textbf{0.88} \\
    \bottomrule
  \end{tabular}
  \caption{
    \textbf{F1-scores of different models for within-subject classification tasks.}
    Each column reports the F1-score for a specific maneuver class, averaged over five trials.
    Bold values highlight ESNNet.
  }
  \label{tab:2}
\end{table}

\subsection{Cross-Subject Classification}
\noindent\textbf{Evaluation.}
We adopt a leave-one-subject-out (LOSO) evaluation strategy: in each fold, data from four subjects are used for training, while the remaining subject is reserved exclusively for testing. No calibration or fine-tuning is performed on the test subject. All other training configurations mirror those in the within-subject setting.

\noindent\textbf{Results.}
As summarized in Table\,\ref{tab:3}, Overall performance declines substantially relative to within-subject classification, with mean accuracies dropping to approximately 50\% across models. ESNNet sustains the highest robustness, achieving an average of 51.3\% $\pm$ 1.4\%, marginally outperforming EEGConformer and FBCNet. However, class imbalance remains evident: most models overpredict the majority pumping class. Subject-level analysis reveals strong individuality effects. e.g., Subject 3 exhibits high training accuracy but severe generalization failure, whereas Subject 2 shows comparatively transferable signal patterns.

As summarized in Table\,\ref{tab:4}, pumping remains the most consistently recognized class, while frontside and backside actions yield poor cross-subject discrimination. Even ESNNet achieves only 0.14 and 0.39 F1 on these two categories, underscoring the persistent generalization bottleneck in EEG decoding.

\begin{table}[htbp]
  \centering
  \setlength{\tabcolsep}{10pt}
  \renewcommand{\arraystretch}{1.3}
  \begin{tabular}{lc}
    \toprule
    Model & Mean Accuracy \\
    \midrule
    EEGNet       & \(50.4 \pm 2.1\)\% \\
    DeepConvNet      & \(49.3 \pm 1.9\)\% \\
    ShallowConvNet   & \(51.1 \pm 1.8\)\% \\
    FBCNet       & \(50.7 \pm 2.0\)\% \\
    EEGConformer & \(51.1 \pm 1.6\)\% \\
    \bfseries\boldmath{ESNNet}  & \bfseries\boldmath{\(51.3 \pm 1.4\)\%} \\
    \bottomrule
  \end{tabular}
  \caption{
    \textbf{Cross-subject evaluation on classification across models.}
    Each entry is the mean $\pm$ standard deviation (std) over five independent runs.
    Bold indicates ESNNet.
  }
  \label{tab:3}
\end{table}

\begin{table}[htbp]
  \centering
  \setlength{\tabcolsep}{4pt}
  \renewcommand{\arraystretch}{1.25}
  \begin{tabular}{lccc}
    \toprule
    Model & Backside (F1) & Frontside (F1) & Pumping (F1) \\
    \midrule
    EEGNet       & 0.08 & 0.07 & 0.41 \\
    DeepConvNet      & 0.10 & 0.00 & 0.53 \\
    ShallowConvNet   & 0.31 & 0.26 & 0.44 \\
    FBCNet       & 0.36 & 0.10 & 0.50 \\
    EEGConformer & 0.24 & 0.06 & 0.52 \\
    \textbf{ESNNet}   & \textbf{0.39} & \textbf{0.14} & \textbf{0.51} \\
    \bottomrule
  \end{tabular}
  \caption{
    \textbf{F1-scores of different models for cross-subject classification tasks.}
    Each column reports the average F1-score for the corresponding maneuver across subjects.
    Bold values highlight ESNNet.%
  }
  \label{tab:4}
\end{table}

\subsection{Understanding ESNNet}
\noindent\textbf{Ablation Study.} 
To assess the contribution of the ESN reservoir, We compare ESNNet with DeepConvNet, a convolution-only baseline of identical front-end design. In the within-subject evaluation, ESNNet attains 83.3\% accuracy, compared to 81.6\% for the baseline. Cross-subject results show a similar pattern, with ESNNet reaching 51.3\% versus 49.3\%. The F1-scores point to more balanced performance, especially in reducing errors on backside and pumping maneuvers. These findings indicate that the reservoir adds incremental value in modeling temporal dependencies.

\noindent\textbf{Visualization.} 
Further work should incorporate visualization methods to examine interpretability. Potential approaches include comparing softmax output distributions, analyzing confusion matrices, and projecting reservoir states with t-SNE. These analyses would provide evidence on whether ESNNet not only improves predictive accuracy but also yields clearer class boundaries and action-specific temporal dynamics.

\section{Conclusion}
We propose ESNNet, a hybrid CNN–ESN architecture for decoding complex motor behaviors from EEG signals. By combining the spatial representation power of CNNs with the temporal modeling efficiency of ESNs, ESNNet provides a computationally lightweight yet effective solution for processing nonstationary, noisy neural signals.

On the multi-class skateboard EEG dataset, ESNNet achieved $83.2\%$ accuracy in within-subject settings and maintained competitive performance under cross-subject evaluation, outperforming or matching state-of-the-art baselines. These results underscore both the promise of reservoir computing in EEG decoding and the enduring challenge of inter-subject variability.

By avoiding costly backpropagation through recurrent layers, ESNNet delivers efficient training and inference, making it well-suited for real-time, resource-constrained deployment. Future research directions include exploring trainable or multi-layer reservoirs, as well as integrating adversarial domain adaptation methods to enhance cross-subject generalization.

Our findings demonstrate that reservoir computing can serve as a robust and interpretable foundation for EEG-based brain–computer interfaces, bridging the gap between individualized performance and generalized applicability.

\end{document}